\documentclass{article}

\usepackage[final]{corl_2019} 

\usepackage{times}

\usepackage[numbers]{natbib}
\usepackage{multicol}
\usepackage{amsmath, amssymb}
\newcommand{\E}{\mathbb{E}}

\usepackage{tikz}
\usepackage{algorithm}
\usepackage{algcompatible}
\usepackage{multicol}
\usepackage{enumitem}
\usepackage{wrapfig}
\usepackage{subcaption}
\usepackage{lipsum}
\usepackage{wrapfig}
\usepackage{array}

\usepackage{graphicx}
\usepackage{adjustbox}
\usepackage[font=small,labelfont=bf]{caption}

\setlist{nolistsep}

\title{Contextual Imagined Goals for Self-Supervised Robotic Learning}

\author{Ashvin Nair$^*$ \hspace{6mm} Shikhar Bahl$^*$ \hspace{2mm} Alexander Khazatsky$^*$ \\ \textbf{Vitchyr Pong \hspace{8mm} Glen Berseth \hspace{8mm} Sergey Levine \hspace{4mm} } \\
\footnotesize{University of California, Berkeley}
}

\begin{document}
\maketitle

\begin{abstract}
While reinforcement learning provides an appealing formalism for learning individual skills, a general-purpose robotic system must be able to master an extensive repertoire of behaviors. Instead of learning a large collection of skills individually, can we instead enable a robot to propose and practice its own behaviors automatically, learning about the affordances and behaviors that it can perform in its environment, such that it can then repurpose this knowledge once a new task is commanded by the user? In this paper, we study this question in the context of self-supervised goal-conditioned reinforcement learning. A central challenge in this learning regime is the problem of goal setting: in order to practice useful skills, the robot must be able to autonomously set goals that are feasible but diverse. When the robot's environment and available objects vary, as they do in most open-world settings, the robot must propose to itself only those goals that it can accomplish in its present setting with the objects that are at hand. Previous work only studies self-supervised goal-conditioned RL in a single-environment setting, where goal proposals come from the robot's past experience or a generative model are sufficient. In more diverse settings, this frequently leads to impossible goals and, as we show experimentally, prevents effective learning. We propose a conditional goal-setting model that aims to propose goals that are feasible from the robot's current state. We demonstrate that this enables self-supervised goal-conditioned off-policy learning with raw image observations in the real world, enabling a robot to manipulate a variety of objects and generalize to new objects that were not seen during training.
\end{abstract}


\section{Introduction}

In order for robots to truly become \emph{generalists}, they must be readily taskable by humans, handle raw sensory inputs without instrumentation, and be equipped with a range of skills that generalize effectively to new situations.
Reinforcement learning autonomously learns policies that maximize a reward function and is a promising approach towards such generalist robots. However, in a general setting involving diverse objects and tasks, prior information about what tasks to learn is hard to come by without manually designing object detectors and reward functions.
How can a robot explore the world in order to learn and fine-tune useful skills on diverse objects, only from acting and observing how its actions affect its sensory stream?

\begin{figure}
    \footnotesize
    \setlength{\unitlength}{\textwidth}





    \begin{picture}(1, 0.6)(0, 0)
        \put(0,0){\includegraphics[width=0.99\textwidth]{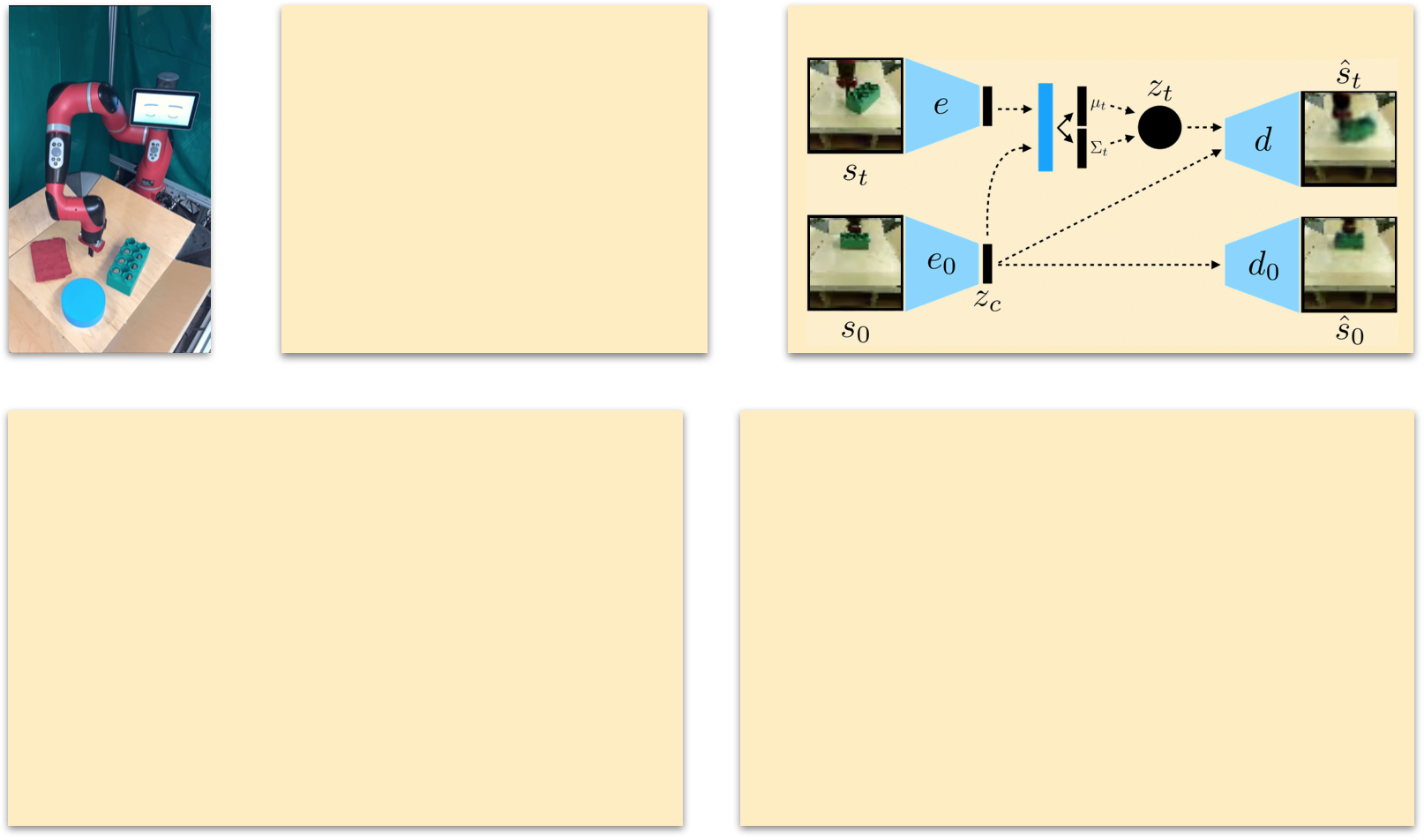}}

        \put(0.15, 0.44){\vector(1, 0){0.08}}

        \put(0.25,0.53){\parbox{1.0in}{1. Collect random interaction samples}}
        \put(0.27, 0.38){\includegraphics[width=0.1\textwidth]{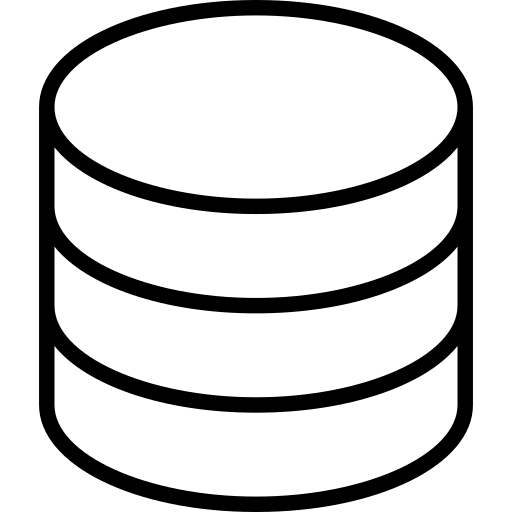}}
        \put(0.25, 0.36){$\{ \tau_1, \tau_2, \dots, \tau_N \}$}
        \put(0.4, 0.44){\vector(1, 0){0.14}}
        \qbezier(0.24, 0.365)(0.16, 0.365)(0.16, 0.31)
        \put(0.16, 0.31){\vector(0, -1){0.01}}

        \put(0.605,0.555){\parbox{2.0in}{2. Train context-conditioned VAE}}
        \qbezier(0.66, 0.35)(0.66, 0.32)(0.6, 0.32)
        \put(0.6, 0.32){\line(-1,0){0.40}}
        \qbezier(0.20, 0.32)(0.18, 0.32)(0.18, 0.31)
        \put(0.18, 0.31){\vector(0, -1){0.01}}

        \put(0.03,0.25){\parbox{2.3in}{3. RL training: learn policy $\pi(\bar{z}_t, \bar{z}_g)$ to minimize latent distance to generated goal $z_g$}}
        \put(0.03,0.11){
            \parbox{2.3in}{

                \hspace{0.04\linewidth} $s_0$ \hspace{0.575\linewidth} $s_H$ \hspace{0.065\linewidth} $d(\bar{z}_g)$

                \includegraphics[width=0.15\linewidth]{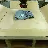}
                \includegraphics[width=0.15\linewidth]{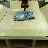}
                \includegraphics[width=0.15\linewidth]{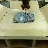}
                \includegraphics[width=0.15\linewidth]{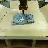}
                \includegraphics[width=0.15\linewidth]{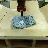}
                \hspace{0.02\linewidth}
                \includegraphics[width=0.15\linewidth]{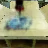}

                \includegraphics[width=0.15\linewidth]{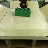}
                \includegraphics[width=0.15\linewidth]{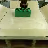}
                \includegraphics[width=0.15\linewidth]{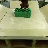}
                \includegraphics[width=0.15\linewidth]{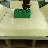}
                \includegraphics[width=0.15\linewidth]{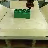}
                \hspace{0.02\linewidth}
                \includegraphics[width=0.15\linewidth]{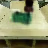}

                \vspace{0.05cm}

                \hspace{0.08\linewidth} Real-world Pusher Training Rollouts
            }
        }

        \put(0.47, 0.14){\vector(1, 0){0.04}}

        \put(0.54,0.25){\parbox{2.3in}{4. Test time: agent executes policy to reach human-provided goal image $s_g$}}
        \put(0.54,0.11){
            \parbox{2.3in}{

                \hspace{0.04\linewidth} $s_0$ \hspace{0.575\linewidth} $s_H$ \hspace{0.1\linewidth} $s_g$

                \includegraphics[width=0.15\linewidth]{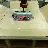}
                \includegraphics[width=0.15\linewidth]{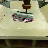}
                \includegraphics[width=0.15\linewidth]{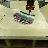}
                \includegraphics[width=0.15\linewidth]{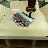}
                \includegraphics[width=0.15\linewidth]{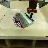}
                \hspace{0.02\linewidth}
                \includegraphics[width=0.15\linewidth]{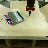}

                \includegraphics[width=0.15\linewidth]{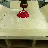}
                \includegraphics[width=0.15\linewidth]{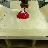}
                \includegraphics[width=0.15\linewidth]{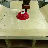}
                \includegraphics[width=0.15\linewidth]{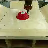}
                \includegraphics[width=0.15\linewidth]{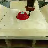}
                \hspace{0.02\linewidth}
                \includegraphics[width=0.15\linewidth]{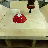}

                \vspace{0.05cm}

                \hspace{0.15\linewidth}
                Test Rollouts (Unseen Objects)
            }
        }
    \end{picture}

    \caption{System overview of our self-supervised learning algorithm.
    (1) The agent collects random interaction data, to be used for both representation learning and as additional off-policy data for RL.
    (2) We propose a context-conditioned generative model (CC-VAE) to learn generalizable skills.
    In order to improve the generation of plausible goal states that result from a starting state $s_0$, our model allows information to flow freely through the context $z_c$ while an information bottleneck on $z_t$ gives us the ability to generate samples by re-sampling only $z_t$.
    This architecture provides a compact representation of the scene disentangling information that changes within a rollout ($z_t$) and information that changes between rollouts ($z_c$).
    We then use $\bar{z}_t = (z_t, z_c)$ as the representation for RL.
    (3) Our proposed CC-RIG algorithm samples latent goals, using the above representation, and learns a policy to minimize the latent distance to the goal with off-policy RL.
    Rollouts are shown on the real-world Sawyer robot pusher environment with visual variation.
    We include the initial image $s_0$, selected frames from the rollout, final image $s_H$, and the decoded goal latent $d(\bar{z}_g)$.
    (4) At test time, the agent is given a goal image $s_g$ and executes the policy to reach it. Our method successfully handles pushing novel objects that were unseen at training time. Example rollouts can be found at  \url{https://ccrig.github.io/}}
    \label{fig:fig1}
\end{figure}

Prior methods have proposed to let an agent learn from its sensor stream by automatically generating plausible goals during an unsupervised training phase, and then learning policies that reach those goals~\cite{nair2018rig, nachum2018hiro, wadefarley2019discern, pong2019skewfit}. Such goals can be defined in a variety of ways, but a simple choice is to use goal observations, such that each proposed task requires reaching a different observation. When the robot observes the world via raw camera images, this corresponds to using images as goals. At test time, a user then provides the robot with a new goal image.

While such methods have been demonstrated in both simulated and real-world settings, they are typically used to learn behaviors in domains with relatively little visual diversity.
In the real world, a robot might interact with highly diverse scenes and objects, and the tasks that it can perform from each of the many possible initial states will be different. If the robot is presented with an object, it can learn to pick up or grasp it, and when it is presented with a door, it can learn to open it. However, it must generate and practice goals that are suitable for each scene. In this paper, we propose and evaluate a self-supervised policy learning method that learns to propose goals that are suitable to the current scene via a conditional goal generation model, allowing it to learn in visually varied settings that prove challenging for prior algorithms.

The key idea in our work is that representing every element in a visually complex scene is often not necessary for control. A scene is a visual form of context that can be factored out, while only the controllable entities in the environment need to be captured for goal setting and representing the state.
To this end, we propose learning a context-conditioned generative model that learns a smooth, compressed latent variable with an information bottleneck, while allowing the context, in the form of the initial state image, to be used freely to reconstruct other images during the task.
This context-conditioned generative model architecture is shown in Figure \ref{fig:fig1}.

The main contribution in this paper builds on this context-conditioned generative model to devise a complete self-supervised goal-conditioned reinforcement learning algorithm, which can handle visual variability in the scene via context-conditioned goal setting. Our method can learn policies that reach visually indicated goals without any additional supervision during training, using the context-conditioned generative model to set goals that are appropriate to the current scene. We show that our approach learns coherent representations of visually varied environments, capturing controllable dimensions of variation while ignoring dimensions that vary but cannot be influenced by the agent, such as lighting and object appearance. We further show that our approach can learn policies to solve tasks in visually varied environments, including in a real-world robotic pushing task with a wide variety of distinct objects.

\section{Related Work}

While many practical robots today perform tasks by executing hand-engineered sequences of motor commands, machine
learning is opening up a new avenue to train a wide variety of robotic tasks from interaction. This body of work includes grasping~\cite{ekvall2004interactive, kroemer2010grasping, bohg2010learning}, and general tasks ~\cite{PETERS2008682,kober2013reinforcement},
multi-task learning~\cite{6907421}, baseball~\cite{peters2008baseball}, ping-pong~\cite{peters2010reps}, and various other tasks \cite{deisenroth2011pilco}. More recently, using expressive function approximators such as neural networks has reduced manual feature engineering and has increased task complexity and diversity, finding use in decision-making domains, such as solving Atari games \cite{mnih2013atari} and Go \cite{silver2016alphago}. Deep learning for robotics has proved to be difficult due to a host of challenges including noisy state estimation, specifying reward functions, and handling continuous action spaces, but has been used to investigate
grasping \cite{pinto2015supersizing}, pushing \cite{agrawal2016poking}, manipulation of 3D object models~\cite{krainin2011autonomous}, active learning~\cite{martinez2014active} and pouring liquids~\cite{schenck2017visual}.
Deep reinforcement learning, which autonomously maximizes a given reward function, has been used to solve precise manipulation tasks \cite{levine2016gps}, grasping \cite{pinto2017robust, levine2017grasping},
door opening \cite{gu2016naf}, and
navigation \cite{kahn2018navigation}.
These methods have succeeded on specific tasks, often with hard-coded reward functions.
However, to scale task generalization robots may need to learn methods that can handle significant environment variation and require relatively little external supervision.

Several works have investigated self-supervised robotic interaction with varied objects in the deep learning setting with the goal of generalizing between objects.
For example, in the domain of robotic grasping, several works have studied autonomous data collection to learn to grasp from a hand-specified grasping reward~\cite{pinto2015supersizing, levine2017grasping}. However, hand-specifying such rewards in general settings and for arbitrary manipulation skills is very cumbersome.
Other work has focused on self-supervised learning with
visual forward models, either by enforcing a simplified dynamical structure \cite{watter2015embed, zhang2019solar} or with pixel transformer architectures \cite{finn2016visualforesight, ebert2017videoprediction, ebert2018retrying, lee2018videoprediction, ebert2018journal}.
However, these methods rely on accurate visual forward modelling, which is itself a very challenging problem.
Instead, we build on self-supervised model-free approaches, which allow the agent to efficiently reach visual goals without planning with a visual forward model.

Prior work has also sought to perform self-supervised learning with model-free approaches. Using visual inverse models \cite{agrawal2016poking} is one such approach, but may not work well for complex interaction dynamics or longer horizon planning.
Most closely related to our approach are prior methods on goal-conditioned reinforcement learning~\cite{kaelbling1993goals, schaul2015uva, andrychowicz2017her}. The methods have been extended to frame self-supervised RL as learning goal reaching with automatically proposed goals, including visually-specified goals~\cite{nair2018rig, pong2019skewfit, wadefarley2019discern, florensa2019self, lin2019rlwithoutstate}. However, they generally focus on learning in narrow environments with little between-trial variability.
In this setting, any previously visited state represents a valid goal. However, in the general case, this is no longer true: when the robot is presented with a different scene or different objects on each trial, it must only set those goals that can be accomplished in the current scene.
In contrast, we focus on enabling self-supervised learning from off-policy data in heterogeneous environments with increased factors of variability.

\section{Background}

In this section, we provide an overview of relevant prior work necessary to understand our method, including goal-conditioned reinforcement learning and representation learning with variational auto-encoders.

\subsection{Goal-Conditioned Reinforcement Learning}
In an MDP consisting of states $s_t \in \mathcal S$, actions $a_t \in \mathcal A$, dynamics $p(s_{t+1}|s_t, a_t)$, rewards $r_t$, horizon $H$, and discount factor $\gamma$, reinforcement learning addresses optimizing a policy $a_t = \pi_\theta(s_t)$ to maximize expected return $\E[\sum_{t=0}^{H} \gamma^t r_t]$. To learn a variety of skills, it is instead convenient to optimize over a family of reward functions parametrized by goals, as in the framework of goal-conditioned RL \cite{kaelbling1993goals}. A variety of RL algorithms exist, but as we are primarily interested in sample-efficient off-policy learning, we consider goal-conditioned Q-learning algorithms \cite{schaul2015uva}. These algorithms learn a parametrized Q-function $Q_w(s, a, g)$ that estimates the expected return of taking action $a$ from state $s$ with goal $g$. Q-learning methods rely on minimizing the Bellman error:
\begin{align} \label{eq:bellman}
    \mathcal{L} = |Q_w(s, a, g) - (r + \max_{a'} Q_w(s', a', g))|
\end{align}
This objective can be optimized using standard actor-critic algorithms using a set of transitions $(s, a, s', g, r)$ which can be collected off-policy \cite{lillicrap2015continuous}.
In practice, a target network is often used for the second Q-function.

\subsection{Variational Auto-Encoders}
Above, the goal description can take many forms. To handle high-dimensional goals, in this work, we learn a latent representation of the state using a variational auto-encoder (VAE).
A VAE is a probabilistic generative model that has been shown to learn structured representations of high-dimensional data \cite{kingma2014vae} successfully. It is trained by reconstructing states $s$ with a parametrized encoder that converts the state into a normal random distribution $q_\phi(z|s)$ with a parametrized decoder that converts the latent variable $z$ into a predicted state distribution $p_\psi(s|z)$, while keeping the latent $z$ close (in KL divergence) to its prior $p(z)$, a standard normal distribution. To train the encoder and decoder parameters $\phi$ and $\psi$, we jointly optimize both objectives when minimizing the negative evidence lower bound:
\begin{align}
    \mathcal L_\text{VAE} = -\E_{q_\phi(z|s)} [\log p(s|z)] + \beta D_{KL}(q_\phi(z|s)||p(z)).
\end{align}

\subsection{Conditional Variational Auto-Encoders}
\label{sec:cvae}

Instead of a generative model that learns to generate the dataset distribution, one might instead desire a more structured generative model that can generate samples based on structured input.
One example of this is a conditional variational auto-encoder (CVAE) that conditions the output on some input variable $c$ and samples from $p(x|c)$ \cite{sohn2015cvae}.
For example, to train a model that generates images of digits given the desired digit, the input variable $c$ might be a one-hot encoded vector of the desired digit.

A CVAE trains $q_\phi(z|s, c)$ and $q_\psi(s|z, c)$, where both the encoder and decoder has access to the input variable $c$. The CVAE then minimizes:
\begin{align} \label{eq:isc-vae-loss}
    \mathcal L_\text{CVAE} = -\E_{q_\phi(z|s, c)} [\log p(s|z, c)] + \beta D_{KL}(q_\phi(z|s, c)||p(z)).
\end{align}
Samples are generated by first sampling a latent $z \sim p(z)$. Based on $c$, we can then decode $z$ with $q_\psi(s|z, c)$ and visualize the output, which is in our case an image.
In our framework $c = s_{0}$.

\subsection{Reinforcement Learning with Imagined Goals}
\label{sec:rig}

To learn skills from raw observations in a self-supervised manner, reinforcement learning with imagined goals (RIG) proposed to use representation learning combined with goal-conditioned RL \cite{nair2018rig}.
The aim of RIG is to choose actions in order to make the state $s_t$ reach a goal image $s_g$ at test time.
RIG first collects an interaction dataset $\{s_t\}$ and learns a latent representation by training a VAE on this data.
Then, a goal-conditioned policy is trained to act in the environment in order to reach a given goal $z_g$.
Exploration data is collected by rolling out the policy with goals that are ``imagined'' from the VAE prior; at test time, the policy can take in a goal image as input, encode the image to the latent space, and act to reach the goal latent.
A key limitation of this method is that sampling goals from the VAE prior during training time assumes that every state in the dataset is reachable at any time.
However, in general, this assumption may not be true.

\section{Self-Supervised Learning with Context-Conditioned Representations}

In this work, our goal is to enable the learning of flexible goal-conditioned policies that can be used to successfully perform a variety of tasks in a variety of contexts -- e.g., with different objects in the scene. Such policies must learn from large amounts of experience, and although it is in principle possible to use random exploration to collect this experience, this quickly becomes impractical in the real world.
It is, therefore, necessary for the robot to set its own goals during self-supervised training, to collect meaningful experience. However, in diverse settings, many randomly generated goals may not be feasible -- e.g., the robot cannot push a red puck if the red puck is not present in the scene.
We propose to extend off-policy goal-conditioned reinforcement learning with a conditional goal setting model, which proposes only those goals that are currently feasible. This enables a learning regime with imagined goals that is more realistic for real-world robotic systems that must generalize effectively to a range of objects and settings.

\subsection{Context-Conditioned VAEs}
\label{sec:ccvae}

To train a generative model that can improve the generation of feasible goals in varied scenes, we use a modified CVAE that uses the initial state $s_{0}$ in a rollout as the input $c$, which we call the ``context" for that rollout.
The modified CVAE, which we call a context-conditioned VAE (CC-VAE), is shown in Figure \ref{fig:fig1}. While most CVAE applications use a one-hot vector as the input, we use an image $s_0$. This image is encoded with a convolutional encoder $e_0$ into a compact representation $z_c$. Note that by design, $e$ and $e_0$ do not share weights, as they are intended to encode different factors of variation in the images. The context $z_c$ is used to output the latent representation $z_t$, as well as the reconstruction of the state $\hat{s}_t$. In addition, $z_c$ is used alone to (deterministically) decode $\hat{s}_0 = d_0(z_c)$. The objective is given by
\begin{align} \label{eq:cc-vae-loss}
    \mathcal L_\text{CC-VAE} = \mathcal L_\text{CVAE} + \log p(s_0|z_c).
\end{align}
Due to the information bottleneck on $z_t$, this loss function penalizes information passing through $z_t$ but allows for unrestricted information flow from $z_c$. Therefore, the optimal solution would encode as much information as possible in $z_c$, while only including the state information that changes within a trajectory in the latent variable $z_t$. These are precisely the features of most interest for control.

\begin{wrapfigure}{r}{0.5\textwidth}
    \begin{minipage}[t]{0.99\linewidth}
        \begin{algorithm}[H]
           	\footnotesize
           	\caption{Context-Conditioned RIG}
           	\label{alg:tbd}
           	\begin{algorithmic}[1]
            \REQUIRE Encoders $\mu(s_t, s_0)$, $e_0(s_0)$, policy $\pi_\theta(\bar{z}, \bar{z_g})$, goal-conditioned value function $Q_w(\bar{z}, \bar{z_g})$, dataset $\mathcal D = \{\tau^{(i)}\}$ of trajectories.
            \STATE Train CC-VAE on $\mathcal D$ by optimizing \eqref{eq:cc-vae-loss}.
            \FOR{$n=0,...,N-1$ episodes}
                \STATE Sample latent goal $z_g \sim p(z)$, $\bar{z}_g = (z_g, z_c)$.
                \STATE Sample initial state $s_0 \sim p(s_0)$.
                \STATE Encode $z_c = e_0(s_0)$
                \FOR{$t=0,...,H -1$ steps}
                    \STATE Observe $s_t$ and encode $\bar{z}_t = (\mu(s_t, s_0), z_c)$
                    \STATE Get action $a_t = \pi_\theta(\bar{z}_t, \bar{z}_g) + \text{noise}$.
                    \STATE Get next state $s_{t+1} \sim p(\cdot \mid s_t, a_t)$.
                    \STATE Store $(\bar{z}_t, a_t, \bar{z}_{t+1}, \bar{z}_g)$ into replay buffer $\mathcal R$.
                    \STATE Sample transition $(\bar{z}, a, \bar{z}', \bar{z}_g) \sim \mathcal R$.
                    \STATE Compute new reward $r = -||\bar{z}' - \bar{z}_g||$.
                    \STATE Minimize \eqref{eq:bellman} using $(\bar{z}, a, \bar{z}', \bar{z}_g, r)$.
                \ENDFOR
            \ENDFOR
           	\end{algorithmic}
        \end{algorithm}
        \vspace{-0.2cm}
    \end{minipage}
\end{wrapfigure}

\subsection{Context-Conditioned Reinforcement Learning with Imagined Goals}
\label{sec:ccrig}

We propose to use our context-conditioned VAE in the RIG framework to learn policies over environments with visual diversity, where each episode might involve interacting with a different scene and different objects.
We first collect a dataset of trajectories $\mathcal D = \{\tau^{(i)}\}$  by executing random actions in the environment. We then learn a CC-VAE, as detailed in Section~\ref{sec:ccvae}, to learn a factored representation of the image observations.
To use the CC-VAE for self-supervised learning, we save the first image $s_0$ when starting a rollout. We compute the encoding of $s_0$, $z_c = e_0(s_0)$. Let $\bar{z}$ denote the context concatenated vector
$(z, z_c)$,
and let $\mu(s, s_0)$ denote the mean of $q_\phi(z|s, s_0)$. We then use RIG in the $\bar{z}$ latent space by encoding observations with $\mu$, meaning that we train a goal-conditioned policy $\pi(\bar{z}, \bar{z_g})$ and a goal-conditioned Q-function $Q(\bar{z}, \bar{z_g})$.

To collect data, we sample a latent goal for each rollout from the prior $z_g \sim N(0, I)$, as in RIG. For every observation $s_t$, we compute the mean encoding $\mu_t(s_t, s_0)$. We then obtain a rollout of the policy by executing $\pi(\bar{z}, \bar{z_g})$. The reward at each timestep is the latent distance $||\bar{\mu}_t - \bar{z}_g||$.

The policy and Q-function can be trained with any off-policy reinforcement learning algorithm. We use TD3 in our implementation \cite{fujimoto2018td3}. Our policy and Q-function are goal-conditioned, and we take advantage of being able to relabel the goals for each transition to improve sample efficiency \cite{andrychowicz2017her, nair2018rig, pong2019skewfit}. However, when relabeling a goal $\bar{z}_g$ with a random goal from the environment, the context-conditioning is still preserved. That is, if $z_g' \sim N(0, 1)$ is the new sampled goal, we use $\bar{z}_g' = (z_g', z_c)$. This ensures that the relabeled goal is compatible with the scene for the corresponding transition.

After training, we can use the learned policy $\pi$ to reach a visually indicated goal.
Given a goal image $s_g$, we encode it into a latent goal $z_g = \mu(s_g, s_0)$.
Then, we execute the policy with the latent goal $\bar{z}_g$, just as during the training phase.
The complete algorithm is presented in Algorithm \ref{alg:tbd}.

\section{Experiments}

In our experiments, we aim to answer the following questions:
\begin{enumerate}
    \item How does our method compare to prior work at learning self-supervised skills in visually diverse environments?
    \item Do context-conditioned VAEs learn an image representation that produces coherent and diverse goals that are suitable for the current scene?
    \item Can our proposed context-conditioned RIG method handle diverse real-world data and learn effective policies under visual variation in the real world?
\end{enumerate}

\begin{figure}[t]
    \centering
    \begin{subfigure}[b]{0.48\textwidth}
        \center
        \includegraphics[height=3.8cm]{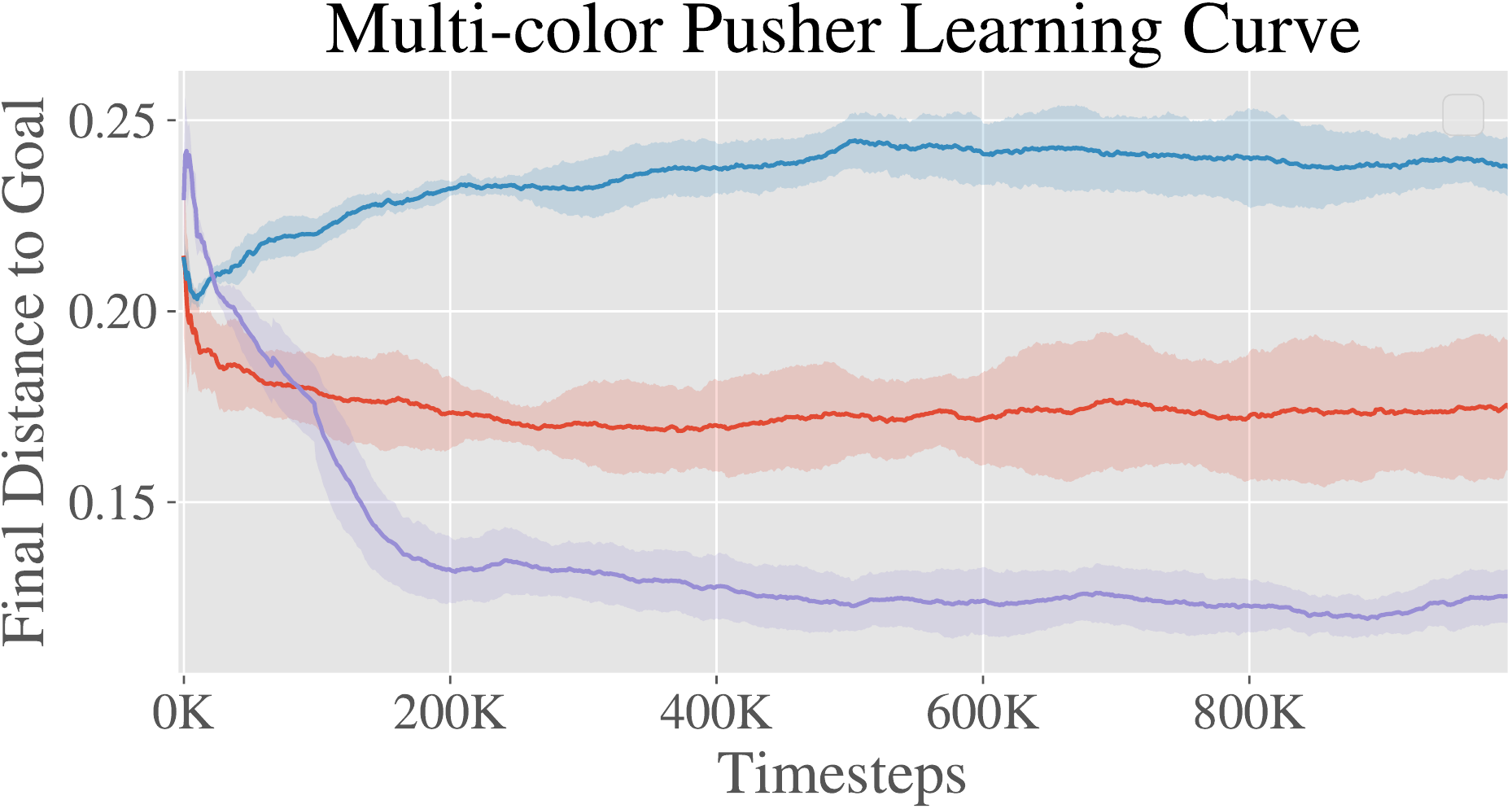}
    \end{subfigure}
    \hspace{0.3cm}
    \begin{subfigure}[b]{0.48\textwidth}
        \includegraphics[height=3.8cm]{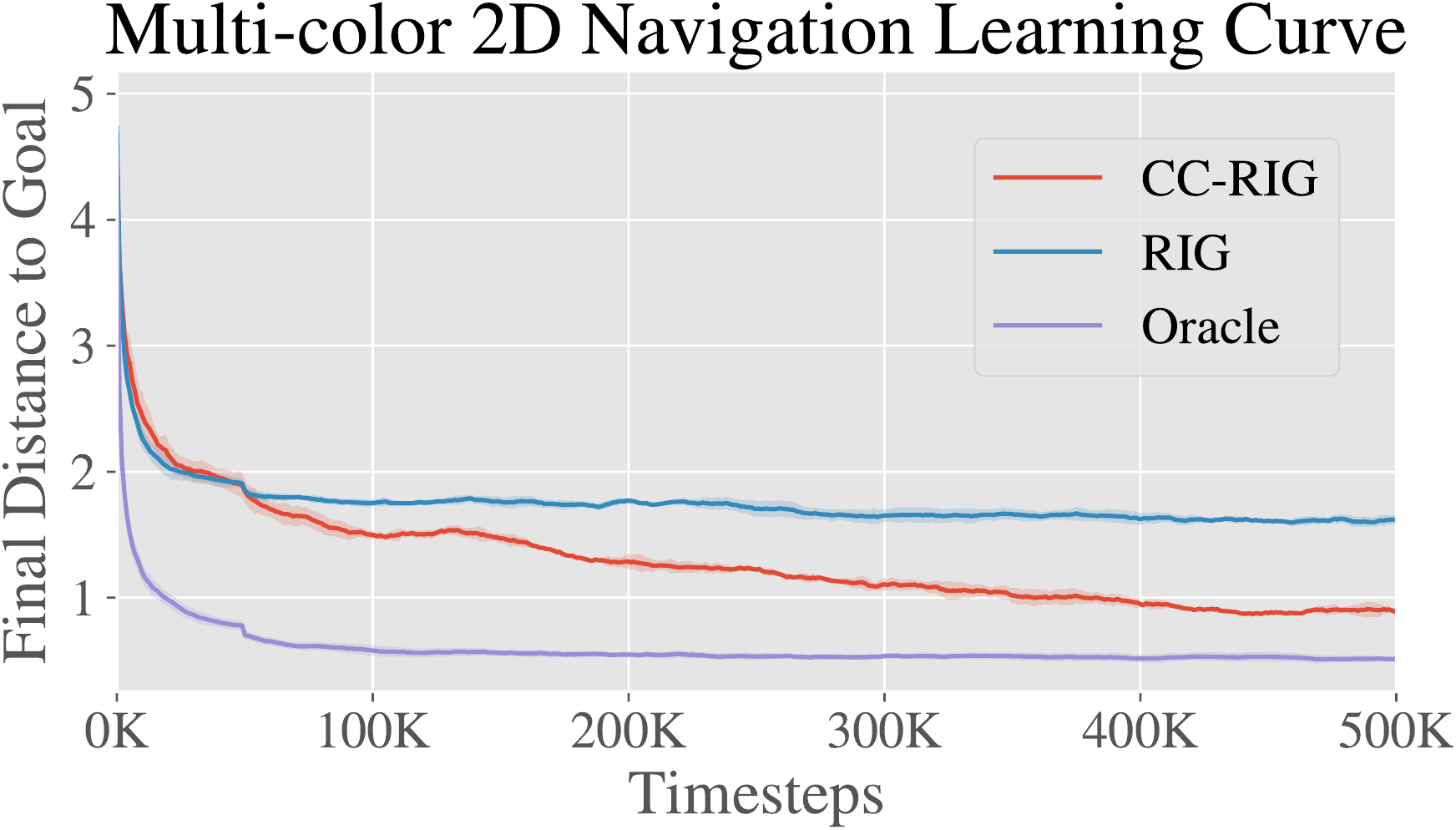}
    \end{subfigure}
    \caption{Self-supervised learning results in visually varied simulated environments. CC-RIG significantly outperforms RIG and is competitive with the oracle method that has direct access to ground truth states. The simulated pusher environment (left) is shown in Figure~\ref{fig:cvae_samples} and the navigation environment is shown in Figure~\ref{fig:pointmass}.}
    \vspace{-0.2in}
    \label{fig:sim-learning-curves}
\end{figure}

\subsection{Self-Supervised Learning in Simulation}

\begin{figure}
    \centering
    \begin{subfigure}[b]{0.16\textwidth}
        \center
        Context $s_0$ \vspace{0.4cm}
    \end{subfigure}
    \begin{subfigure}[b]{0.4\textwidth}
        \center
        \adjustbox{trim=0 0 100 0,clip} {
            \rotatebox[origin=c]{180}{
            \includegraphics[height=31.5px]{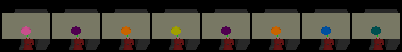}
            }
        }
    \end{subfigure}
    \hspace{0.1cm}
    \begin{subfigure}[b]{0.4\textwidth}
        \center
        \adjustbox{trim={0.39\height} {0.875\height} 0 0,clip} {
        \includegraphics[height=252.5px]{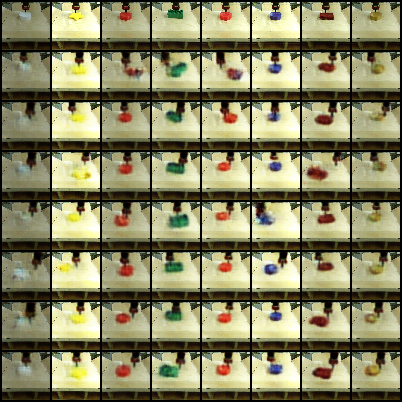}
        }
    \end{subfigure}
    \vspace{0.1cm}

    \begin{subfigure}[b]{0.16\textwidth}
        \center
        CVAE samples \vspace{1.5cm}
    \end{subfigure}
    \begin{subfigure}[b]{0.4\textwidth}
        \center
        \adjustbox{trim=0 0 100 0,clip} {
            \rotatebox[origin=c]{180}{
            \includegraphics[height=92.4px]{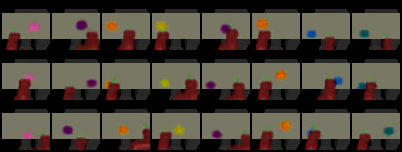}
            }
        }
    \end{subfigure}
    \hspace{0.1cm}
    \begin{subfigure}[b]{0.4\textwidth}
        \center
        \adjustbox{trim={0.39\height} {.125\height} 0 {.5\height},clip} {
        \includegraphics[height=252.5px]{img/cvae_samples_real_pusher3b.png}
        }
    \end{subfigure}
    \vspace{0.1cm}

    \begin{subfigure}[b]{0.16\textwidth}
        \center
        VAE samples \vspace{0.3cm}
    \end{subfigure}
    \begin{subfigure}[b]{0.4\textwidth}
        \center
        \adjustbox{trim=2.2 33.5 100 31.8,clip} {
            \rotatebox[origin=c]{180}{
            \includegraphics[height=92.4px]{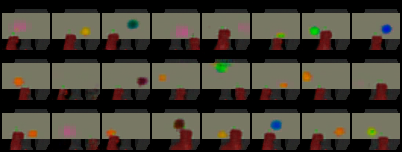}
            }
        }
    \end{subfigure}
    \hspace{0.1cm}
    \begin{subfigure}[b]{0.4\textwidth}
        \hspace{-5px}
        \adjustbox{trim=0 {.26\height} {0.39\height} {.62\height},clip} {
        \includegraphics[height=252.5px]{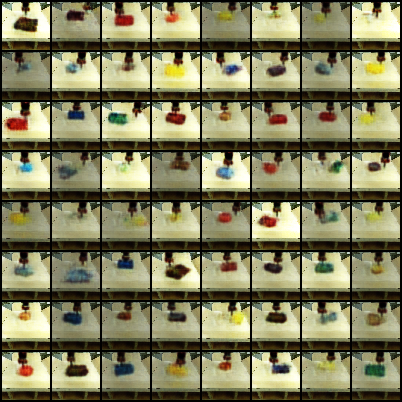}
        }
    \end{subfigure}

    \caption{Comparing samples from our CC-VAE model to a standard VAE. The initial image $s_0$ is shown on the top row, and samples conditioned on $s_0$ are shown below. Our model coherently maintains object color and geometry in its samples, suggesting that the context conditioned model can successfully factor out the scene-specific object identity from the variable object position. This enables the use of the CC-VAE for goal proposals in visually diverse scenes.}
    \vspace{-0.2in}
    \label{fig:cvae_samples}
\end{figure}

In simulation, we can conduct controlled experiments and evaluate against known underlying state values to measure the performance of our approach and prior methods. As a simulation test-bed, we use a multi-color pusher environment simulated in MuJoCo~\cite{todorov12mujoco}. In this environment (shown on the left in Figure~\ref{fig:cvae_samples}),
a simulated Sawyer arm is tasked with pushing a circular puck to a target position, specified by a goal image at test time. On each rollout, the puck color is set to a random RGB value. Therefore, the goal proposals for each method must adequately account for the color of the puck -- a goal that requires moving a red puck to a given location is impossible if only a blue puck is present in the scene.

We compare the following algorithms:
\textbf{CC-RIG.} Our method using a CC-VAE for representation learning, as described in Section~\ref{sec:ccrig}.
\textbf{RIG.} Reinforcement learning with imagined goals \cite{nair2018rig} using a standard VAE, as described in Section \ref{sec:rig}.
\textbf{Oracle.} The oracle agent runs goal-conditioned RL with direct access to state information. Achieving performance similar to the oracle indicates that an algorithm loses little from using raw image observations over ground truth state.

Learning curves comparing these methods are presented in the plot on the left in~Figure~\ref{fig:sim-learning-curves}. CC-RIG outperforms RIG significantly, and standard RIG is not able to improve beyond the initial random policy. The performance of CC-RIG approaches that of the oracle policy, which has access to the true state. This suggests that, in visually varied environments, self-supervised learning is possible so long as the visual complexity is factored out with representation learning, and the proposed goals are consistent with the appearance of the current scene.

\subsection{Generalizing to Varying Appearance and Dynamics with Self-Supervised Learning}

In this experiment, to study changing both visual appearance and physical dynamics, we study how well our method can generalize when the environment dynamics change. We use a simulated 2D navigation task, where the goal is to navigate a point robot around an obstacle. The arrangement of the obstacles is chosen from a set of 15 possible configurations, and the color of the point robot is generated from a random RGB value. Learning curves obtained by training the different methods above in this environment are presented in~Figure~\ref{fig:sim-learning-curves}.
CC-RIG requires more samples to learn, but eventually approaches the oracle performance. RIG, in comparison, plateaus with poor performance. This environment is explained further in the supplementary, in Section~\ref{sec:pointmass} and Figure~\ref{fig:pointmass}.

\subsection{Context-Conditioned VAE Goal Sampling}

To better understand why CC-RIG outperforms RIG, we compare the samples from our CC-VAE to a standard VAE. Samples from both models are shown in Figure~\ref{fig:cvae_samples}.
The quality of the samples reveals why the CC-VAE provides better goal setting for self-supervised learning.
In all environments, the samples from the CC-VAE maintain the background, object shape, and object color from the initial state. Therefore, the goals are more meaningful in the CC-VAE latent space.

This kind of visualization is a good indicator for the suitability of the representation for self-supervised learning.
Diverse, coherent samples indicate that the latent space captures the appropriate factors of change in the environment and can be useful for self-supervised policy learning.
Good samples also suggest that the latent space is well-structured, and therefore distances in the latent space should provide a good reward function for goal-reaching. In practice, we also look at the quality of the reconstructions. Good reconstructions confirm that the latent variables capture sufficient information about the image to be used in place of the image itself as a state representation.

\subsection{Real-World Robotic Evaluation}
\label{sec:realworldexps}

In this experiment, we evaluate whether our method can handle manipulating visually varied objects in the real world. We use CC-RIG to train a Sawyer robot to manipulate a variety of objects, placed one at a time in a bin in front of the robot. As before, the training phase is self-supervised, and the robot must match a given goal image at test time. The robot setup is shown in Figure \ref{fig:fig1}.

\begin{figure}
    \renewcommand{\arraystretch}{1.5}
    \setlength{\tabcolsep}{5pt}
    \newcolumntype{P}[1]{>{\centering\arraybackslash}p{#1}}
    \begin{subfigure}[b]{0.99\textwidth}
        \center
        Real-World Pushing Results
        \begin{tabular}{ | p{2.2cm} | P{2.3cm}| P{2.0cm} | P{2.0cm} | P{3cm} | } \hline
         & CVAE distance & VAE distance & Pixel distance & Object distance (cm) \\ \hline
RIG & $2.37$ $\pm$ $0.97$ & $2.41$ $\pm$ $0.93$ & $93.9$ $\pm$ $41.7$ & $17.1$ $\pm$ $8.2$ \\ \hline
CC-RIG & $1.66$ $\pm$ $0.63$ & $2.17$ $\pm$ $0.88$ & $56.8$ $\pm$ $34.5$ & $14.0$ $\pm$ $6.9$ \\ \hline
CC-RIG, novel & $1.51$ $\pm$ $0.71$ & $1.91$ $\pm$ $0.87$ & $53.1$ $\pm$ $24.9$ & $11.5$ $\pm$ $2.9$ \\ \hline
        \end{tabular}
    \end{subfigure}
    \vspace{0.2cm}

    \begin{subfigure}[b]{0.49\textwidth}
        \center
        Test rollouts, training objects \\
        $s_0$ \hspace{3.7cm} $s_H$ \hspace{0.5cm} $s_g$

        \includegraphics[width=0.14\linewidth]{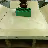}
        \includegraphics[width=0.14\linewidth]{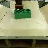}
        \includegraphics[width=0.14\linewidth]{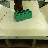}
        \includegraphics[width=0.14\linewidth]{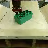}
        \includegraphics[width=0.14\linewidth]{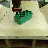}
        \hspace{0.01\linewidth}
        \includegraphics[width=0.14\linewidth]{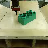}

        \includegraphics[width=0.14\linewidth]{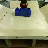}
        \includegraphics[width=0.14\linewidth]{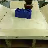}
        \includegraphics[width=0.14\linewidth]{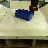}
        \includegraphics[width=0.14\linewidth]{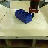}
        \includegraphics[width=0.14\linewidth]{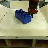}
        \hspace{0.01\linewidth}
        \includegraphics[width=0.14\linewidth]{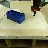}
    \end{subfigure}
    \hfill \rule[4pt]{1pt}{2cm} \hfill
    \begin{subfigure}[b]{0.49\textwidth}
        \center
        Test rollouts, novel objects \\
        $s_0$ \hspace{3.7cm} $s_H$ \hspace{0.5cm} $s_g$

        \includegraphics[width=0.14\linewidth]{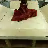}
        \includegraphics[width=0.14\linewidth]{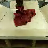}
        \includegraphics[width=0.14\linewidth]{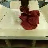}
        \includegraphics[width=0.14\linewidth]{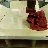}
        \includegraphics[width=0.14\linewidth]{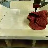}
        \hspace{0.01\linewidth}
        \includegraphics[width=0.14\linewidth]{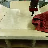}

        \includegraphics[width=0.14\linewidth]{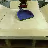}
        \includegraphics[width=0.14\linewidth]{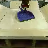}
        \includegraphics[width=0.14\linewidth]{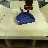}
        \includegraphics[width=0.14\linewidth]{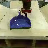}
        \includegraphics[width=0.14\linewidth]{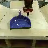}
        \hspace{0.01\linewidth}
        \includegraphics[width=0.14\linewidth]{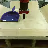}
    \end{subfigure}

    \caption{The table above shows the performance of our method in the real-world, evaluated with four different evaluation metrics\protect\footnotemark. CC-RIG outperforms RIG in each one, even when tested on novel objects that it has not been trained on.
    Test rollouts of our method are shown on training objects on the left and unseen novel objects on the right.
    Successful rollouts where the object is pushed to the goal location are shown in top row, and failure modes are shown in the bottom row. }
    \label{fig:realworld-robot-pushing-results}
\end{figure}

We first collect a large dataset with random actions and train a CC-VAE on the data. Samples from the model are shown in Figure~\ref{fig:realworld-robot-pushing-results}.
The CC-VAE learns to generate goals with the correct object. To handle varying brightness at different times of the day, we added data augmentation by applying a color jitter filter to $(s_0, s_t)$ pairs.
As seen in the figure, the model is robust to this factor of variation. Each sample contains the same type of object, brightness level, and background as the initial state that it is conditioned on. However, crucially, these factors of variation are not present in $z_t$, as evidenced by the fact they do not vary within each column of Figure~\ref{fig:realworld-robot-pushing-results}, but the object position does.

Next, we run CC-RIG with the trained CC-VAE to learn to reach visually indicated goals in a self-supervised manner. We first conduct fully off-policy training using the same dataset as was used to train the CC-VAE, consisting of 50,000 samples (about 3 hours) of total interaction with 20 objects. Then, we collect a small amount of additional on-policy data to finetune the policy, analogous to recent work on large-scale vision-based robotic reinforcement learning~\cite{pmlr-v87-kalashnikov18a}.
The robot learns to push objects to target locations, indicated by a goal image.
The real-world results are presented in Figure~\ref{fig:realworld-robot-pushing-results}.
Because it is difficult to automatically detect the positions of objects, we show some representative rollout examples, and we compute several distance metrics between the final state of a rollout and the goal: \textbf{CVAE distance.} CC-VAE latent space distance between final image and goal. \textbf{VAE distance.} VAE latent space distance between final image and goal. \textbf{Pixel distance.} We manually label the center of mass of the object in the final image and goal image, and compute the distance between them. \textbf{Object distance.} We measure the distance between the physical goal position of the object and the final position. In each metric, CC-RIG outperforms RIG.

At training time, the dataset consists of interaction with 20 objects. The result of running CC-RIG on novel objects that were not included in the dataset are shown in the table as ``CC-RIG, novel'' and in the rollouts in Figure~\ref{fig:realworld-robot-pushing-results}. These results show that our method can also generalize its experience to push novel objects it has not seen before.

\section{Conclusion}
\label{sec:conclusion}

\footnotetext{The first three metrics are computed on 40 trajectories per method, and we report mean $\pm$ standard deviation. Object distance is computed on 10 trajectories per method, and we report median $\pm$ standard deviation.}

We presented a method for sample-efficient, flexible self-supervised task learning for environments with visual diversity. Our method can learn effective behavior without external supervision in simulated environments with randomized colors and layout, and in a real-world pushing task with differently colored pucks. Each environment contains an axis of visual variation that requires our algorithm to utilize an intelligent goal-setting strategy, to ensure that the self-proposed goals are consistent with the tasks and feasible in the current scene.

The main idea behind our method is to devise a context-conditioned goal proposal mechanism, allowing our self-supervised reinforcement learning algorithm to propose goals for itself that are feasible to reach. This context-conditioned VAE model factors out the unchanging context of a rollout, such as which objects are present in the scene, from the controllable aspects, such as the object positions to construct a more generalizable goal proposal model.

We believe this contribution will enable scalable learning in the real world. An agent manipulating objects in the real world must handle many forms of variation: different manipulation skills to learn, objects to manipulate, as well as variation in lighting, textures, etc. Methods that learn from data must be able to represent these variations while at the same time taking advantage of common structure across objects and tasks in order to achieve practical sample efficiency. Future work will address the remaining challenges to achieve this vision.

\acknowledgments{This research was supported in part by the National Science Foundation under IIS-1651843, IIS-1700697, and IIS-1700696, the Office of Naval Research, ARL DCIST CRA W911NF-17-2-0181, DARPA, Berkeley DeepDrive, Google, Amazon, and NVIDIA.}

{ \small
\bibliographystyle{corlabbrvnat}
\bibliography{references}
}

\pagebreak

\noindent\makebox[\linewidth]{\rule{\linewidth}{3.0pt}}
\begin{center}
\LARGE{\textbf{Supplementary Material}}
\end{center}
\noindent\makebox[\linewidth]{\rule{\linewidth}{0.8pt}}

\begin{figure}[b]
    \centering
    \vspace{-0.5cm}
    \begin{subfigure}[b]{0.49\textwidth}
        \begin{subfigure}[b]{0.4\textwidth}
            \center
            Context $s_0$ \vspace{0.2cm}
        \end{subfigure}
        \begin{subfigure}[b]{0.59\textwidth}
            \adjustbox{trim=0 0 {0.5\width} 0,clip} {
                \includegraphics[width=1.7\linewidth]{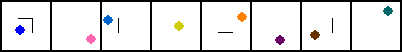}
            }
        \end{subfigure}

        \vspace{0.1cm}

        \begin{subfigure}[b]{0.4\textwidth}
            \center
            CVAE samples \vspace{1cm}
        \end{subfigure}
        \begin{subfigure}[b]{0.59\textwidth}
            \adjustbox{trim=0 0 {0.5\width} 0,clip} {
                \includegraphics[width=1.7\linewidth]{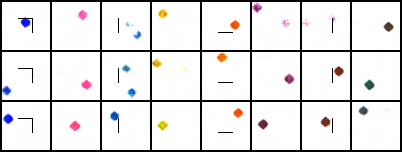}
            }
        \end{subfigure}

        \vspace{0.1cm}

        \begin{subfigure}[b]{0.4\textwidth}
            \center
            VAE samples \vspace{1cm}
        \end{subfigure}
        \begin{subfigure}[b]{0.59\textwidth}
            \adjustbox{trim=0 0 {0.5\width} 0,clip} {
                \includegraphics[width=1.7\linewidth]{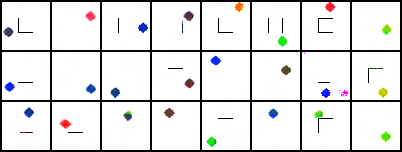}
            }
        \end{subfigure}
    \end{subfigure}
    \hfill
    \begin{subfigure}[b]{0.49\textwidth}
        \begin{center}
            Multi-color 2D Navigation Training Rollouts \vspace{0.1cm}
        \end{center}

        \hspace{0.2cm} $s_0$ \hspace{3.7cm} $s_H$ \hspace{0.7cm} $d(\bar{z}_g)$

        \includegraphics[width=0.14\linewidth]{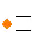}
        \includegraphics[width=0.14\linewidth]{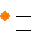}
        \includegraphics[width=0.14\linewidth]{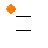}
        \includegraphics[width=0.14\linewidth]{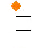}
        \includegraphics[width=0.14\linewidth]{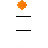}
        \hspace{0.3cm}
        \includegraphics[width=0.14\linewidth]{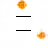}

        \includegraphics[width=0.14\linewidth]{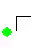}
        \includegraphics[width=0.14\linewidth]{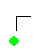}
        \includegraphics[width=0.14\linewidth]{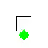}
        \includegraphics[width=0.14\linewidth]{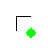}
        \includegraphics[width=0.14\linewidth]{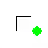}
        \hspace{0.3cm}
        \includegraphics[width=0.14\linewidth]{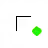}

        \begin{center}
            Multi-color 2D Navigation Test Rollouts \vspace{0.1cm}
        \end{center}

        \hspace{0.2cm} $s_0$ \hspace{3.7cm} $s_H$ \hspace{0.8cm} $s_g$

        \includegraphics[width=0.14\linewidth]{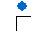}
        \includegraphics[width=0.14\linewidth]{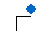}
        \includegraphics[width=0.14\linewidth]{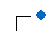}
        \includegraphics[width=0.14\linewidth]{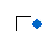}
        \includegraphics[width=0.14\linewidth]{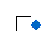}
        \hspace{0.3cm}
        \includegraphics[width=0.14\linewidth]{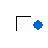}

        \includegraphics[width=0.14\linewidth]{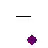}
        \includegraphics[width=0.14\linewidth]{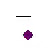}
        \includegraphics[width=0.14\linewidth]{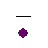}
        \includegraphics[width=0.14\linewidth]{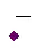}
        \includegraphics[width=0.14\linewidth]{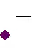}
        \hspace{0.3cm}
        \includegraphics[width=0.14\linewidth]{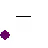}
    \end{subfigure}

    \caption{The pointmass environment is shown. Left, we compare samples from our CC-VAE model to a standard VAE. The initial image $s_0$ is shown on the top row, and samples conditioned on $s_0$ are shown below. Our model coherently maintains object color and geometry in its samples, suggesting that the context conditioned model can successfully factor out the scene-specific object identity from the variable object position. This enables the use of the CC-VAE for goal proposals in visually diverse scenes. Right, rollouts from CC-RIG are shown. We see that during collecting training rollouts, the policy succeeds in coherent exploration by generating reachable goals. Then, given a goal image at test time, the policy successfully reaches the goal.
    Videos of rollouts on all environments, both simulation and real-world, can be found at  \url{https://ccrig.github.io/}
    }
    \label{fig:pointmass}
\end{figure}

\section{Multi-Color 2D Navigation Experiments} \label{sec:pointmass}

In order to study generalizing to varying appearance and dynamics with CC-RIG, we introduced the multi-color 2D point navigation environment shown in \ref{fig:pointmass}. The goal is to navigate a point robot around the central walls. The arrangement of the walls is randomly chosen from a set of 15 possible configurations in each rollout, and the color of the circle indicating the position of the point robot is generated from a random RGB value. Thus at test time, the agent sees new colors it has never trained on.

First, we see from the samples in~Figure~\ref{fig:pointmass} that the learned latent space for the CC-VAE is more reasonable than a VAE: it preserves color and wall information in samples and represents only the colored circle position in the latent variable $z_t$. This improves the capability of our algorithm to learn in several ways: it provides a more informative reward function, and gives us better goal sampling for both exploration rollouts and experience relabeling when training the Q function.

Learning curves obtained by training the different methods above in this environment are presented in the main paper Figure~\ref{fig:sim-learning-curves}. This task is trivial for the oracle method to learn, as it directly receives state information and does not need to generalize between different object appearances. CC-RIG requires more samples to learn, but eventually approaches the oracle performance. RIG plateaus with poor performance in comparison.

\pagebreak

\section{Off-Policy Experiments}

Because we use off-policy RL methods, one major benefit is that we can bootstrap training from large interaction datasets rather than requiring on on-policy data collection. This is particularly vital in the real-world, where on-policy data collection is expensive in terms of human effort, and repeatedly tuning on-policy methods for complex tasks is likely to be impractical. Our robot experiments are therefore run by starting with a fixed initial dataset of 50,000 samples (about 3 hours) of random interaction with 20 objects, which is used for both training the CC-VAE as well as RL. Our simulated experiments are conducted with online data-collection to make comparison with prior work clearer, but in this section we show that bootstrapping with off-policy training is possible in these settings as well.

\begin{figure}[t]
    \centering
    \begin{subfigure}[b]{0.48\textwidth}
        \center
        \includegraphics[height=3.8cm]{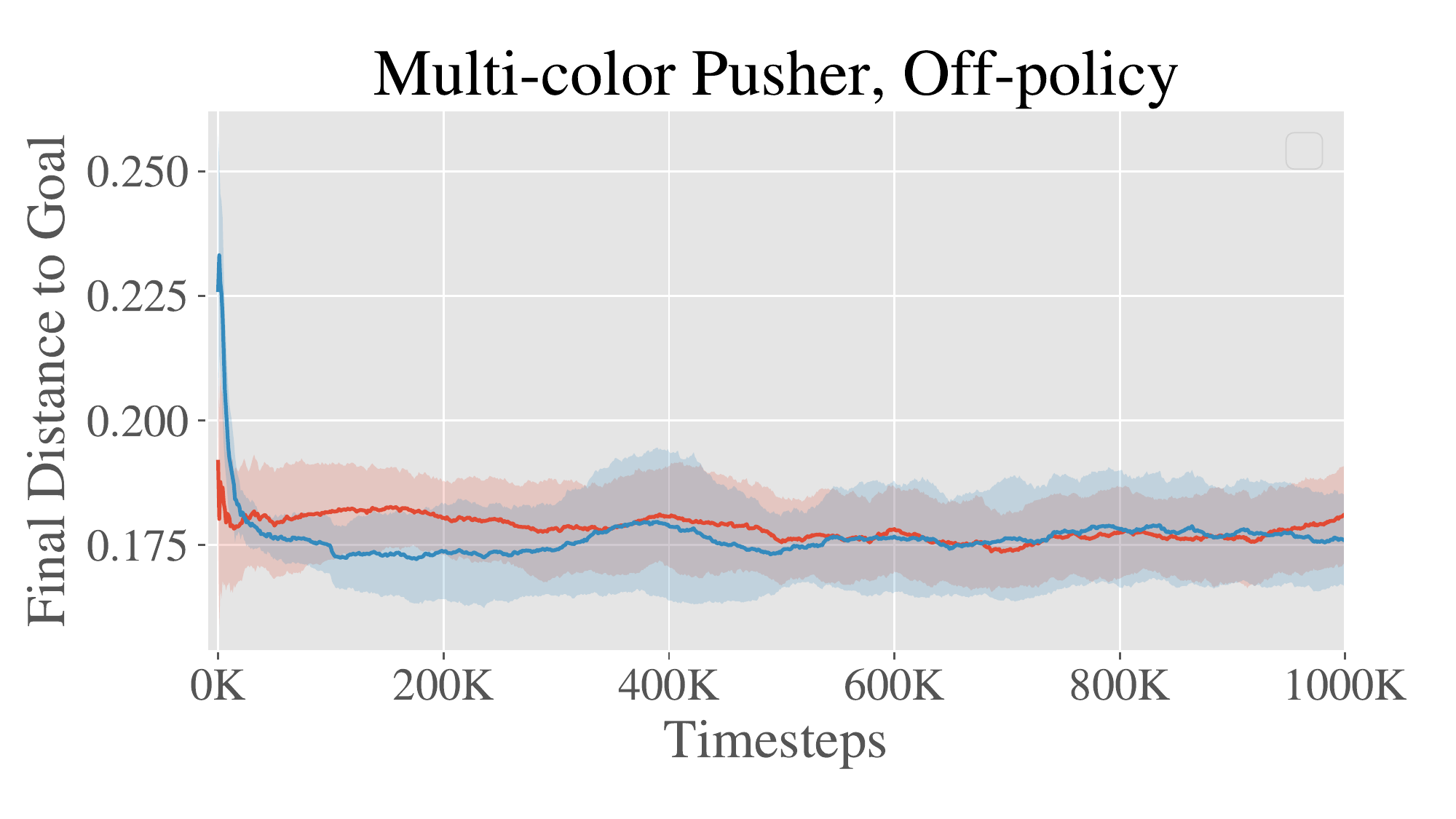}
    \end{subfigure}
    \hspace{0.3cm}
    \begin{subfigure}[b]{0.48\textwidth}
        \includegraphics[height=3.8cm]{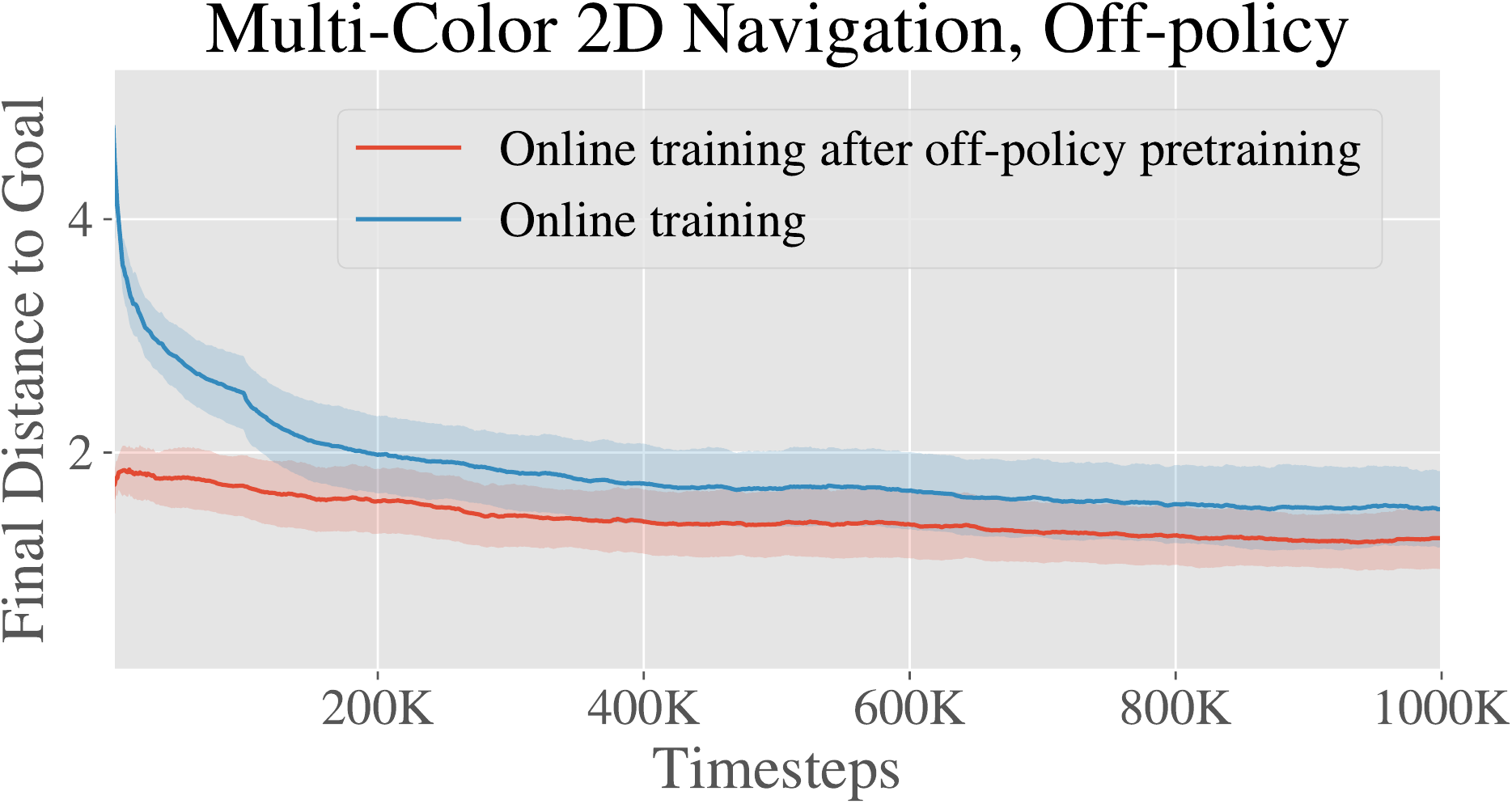}
    \end{subfigure}
    \caption{Off-policy learning results in simulated environments. These experiments begin with 100K samples from random interaction loaded into the replay buffer. The red lines (off-policy pre-training) additionally do 100K steps of Q learning before starting on-line data collection. We see that off-policy pre-training results in proficient initial performance from a fixed dataset, which is extremely useful in domains such as robotics where collecting new samples is expensive.}
    \label{fig:offpolicy}
\end{figure}

In our simulated experiments, we first collect 100,000 samples (1000 trajectories) with random actions. This data is used both to train the CC-VAE and as off-policy data. When we begin RL, we load these samples into the replay buffer and perform 100,000 gradient updates of RL. As shown in Figure \ref{fig:offpolicy}, this allows us to begin online data collection with a reasonably good policy. But we see that online data collection does improve slightly beyond this initial policy. In dynamically sensitive environments or environments where random actions do not provide meaningful interaction, this online data collection may still be very valuable.

\end{document}